\documentclass[preprint,12pt]{elsarticle}

\usepackage{amssymb}
\usepackage{amsmath}
\usepackage{amsmath,amsfonts}
\usepackage{algorithmic}
\usepackage{algorithm}
\usepackage{array}
\usepackage[caption=false,font=footnotesize,labelfont=rm,textfont=rm]{subfig}
\usepackage{textcomp}
\usepackage{makecell}
\usepackage{stfloats}
\usepackage{url}
\usepackage{verbatim}
\usepackage{graphicx}
\usepackage{colortbl} 
\usepackage{hyperref} 
\usepackage{bigstrut}
\usepackage{caption}
\usepackage{amsmath}
\usepackage{multirow}                 % 一般用以设计表格，将所行合并
\usepackage{multicol}                 % 合并多列
\usepackage{multirow} 
\usepackage{siunitx}
\usepackage{pifont}       % \ding{xx}
\usepackage{bbding}    
\usepackage{flushend}

\usepackage{algorithm}
\usepackage{algorithmic}
\usepackage{booktabs,makecell, multirow, tabularx}
\hypersetup{
    colorlinks=true, % 设置链接的颜色，可以设为true或false
    linkcolor=blue, % 设置链接颜色
    citecolor=green, % 设置引用颜色
    urlcolor=blue % 设置URL颜色
}
\journal{Pattern Recognition}

\begin{document}

\begin{frontmatter}



% \title{Visual and Memory Dual Adapter for Multi-Modal Object Tracking}
\title{Learning Frequency and Memory-Aware Prompts for Multi-Modal Object Tracking}

\author[label1]{Boyue Xu} %% Author name
\author[label1]{Ruichao Hou}
\author[label1]{Tongwei Ren}
\author[label2]{Dongming Zhou}
\author[label1]{Gangshan Wu}
\author[label3,label4]{Jinde Cao}
\affiliation[label1]{organization={State Key Laboratory for Novel Software Technology},%Department and Organization
            addressline={Nanjing University}, 
            city={Nanjing},
            postcode={210008}, 
            state={Jiangsu},
            country={China}}
\affiliation[label2]{organization={School of Information Science and Engineering},
            addressline={Yunnan University}, 
            city={Kunming},
            postcode={650091}, 
            state={Yunnan},
            country={China}}
\affiliation[label3]{organization={School of Mathematics},
            addressline={Southeast University}, 
            city={Nanjing},
            postcode={210096}, 
            state={Jiangsu},
            country={China}}
\affiliation[label4]{organization={Purple Mountain Laboratories},
            % addressline={Southeast University}, 
            city={Nanjing},
            postcode={211111}, 
            state={Jiangsu},
            country={China}}
%% Abstract

% \begin{abstract}
% Prompt-learning-based multi-modal trackers have achieved promising progress by employing lightweight visual adapters to incorporate auxiliary modality features into frozen foundation models. However, existing approaches often struggle to learn reliable prompts due to limited exploitation of critical cues across frequency and temporal domains. In this paper, we propose a novel visual and memory dual adapter (VMDA) to construct more robust and discriminative representations for multi-modal tracking. Specifically, we develop a simple but effective visual adapter that adaptively transfers discriminative cues from auxiliary modality to dominant modality by jointly modeling the frequency, spatial, and channel-wise features. Additionally, we design the memory adapter inspired by the human memory mechanism, which stores global temporal cues and performs dynamic update and retrieval operations to ensure the consistent propagation of reliable temporal information across video sequences. Extensive experiments demonstrate that our method achieves state-of-the-art performance on the various multi-modal tracking tasks, including RGB-Thermal, RGB-Depth, and RGB-Event tracking. Code and models are available at \href{https://github.com/xuboyue1999/mmtrack.git}{https://github.com/xuboyue1999/mmtrack.git}.
% \end{abstract}
\begin{abstract}
Prompt-learning–based multi-modal trackers have made strong progress by using lightweight visual adapters to inject auxiliary-modality cues into frozen foundation models. However, they still underutilize two essentials: modality-specific frequency structure and long-range temporal dependencies. We present Learning Frequency and Memory-Aware Prompts, a dual-adapter framework that injects lightweight prompts into a frozen RGB tracker. A frequency-guided visual adapter adaptively transfers complementary cues across modalities by jointly calibrating spatial, channel, and frequency components, narrowing the modality gap without full fine-tuning. A multilevel memory adapter with short, long, and permanent memory stores, updates, and retrieves reliable temporal context, enabling consistent propagation across frames and robust recovery from occlusion, motion blur, and illumination changes. This unified design preserves the efficiency of prompt learning while strengthening cross-modal interaction and temporal coherence. Extensive experiments on RGB-Thermal, RGB-Depth, and RGB-Event benchmarks show consistent state-of-the-art results over fully fine-tuned and adapter-based baselines, together with favorable parameter efficiency and runtime. Code and models are available at \href{https://github.com/xuboyue1999/mmtrack.git}{https://github.com/xuboyue1999/mmtrack.git}.
\end{abstract}


% \begin{highlights}
% \end{highlights}

%% Keywords

\end{frontmatter}




\section{Introduction}
\begin{figure}[t]
\centering
  \includegraphics[width=0.7\textwidth]{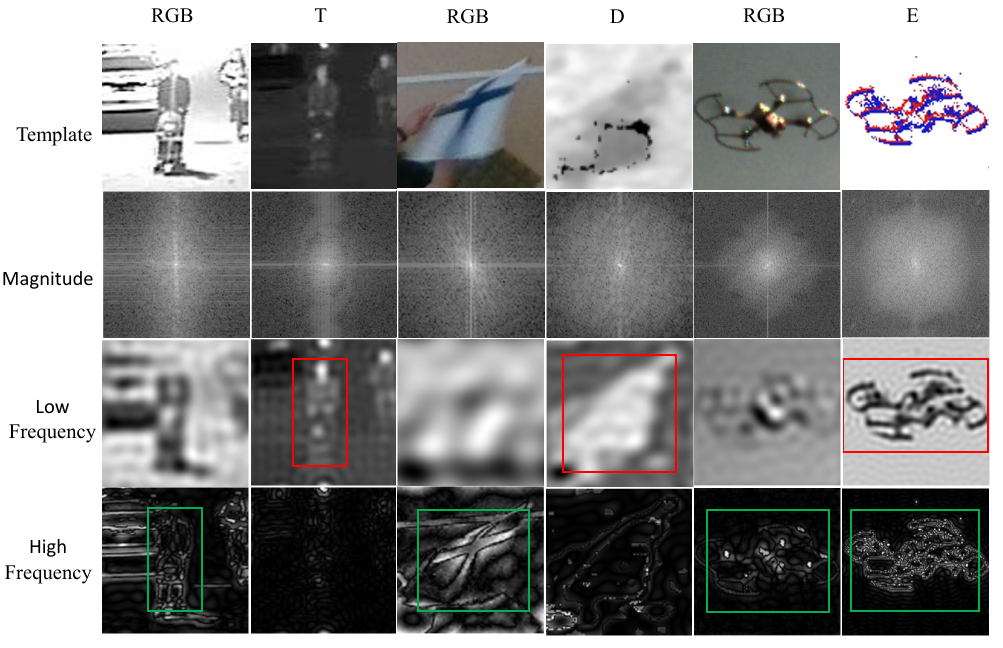}   
  \caption{Illustration of frequency-domain characteristics for RGB-T, RGB-D and RGB-E. 
  The second to fourth rows show the magnitude map, the low-frequency visualization, and the high-frequency visualization, respectively. 
  Red boxes indicate the most informative regions in the low-frequency domain, while green boxes highlight the most informative regions in the high-frequency domain.}   
  \label{fig:intro2}  
        % \vspace{-15pt}
\end{figure}

Visual object tracking (VOT)~\cite{tra1,tra2,tra5,tra6} aims to localize the target annotated in the first frame throughout subsequent frames~\cite{rgbtREVIEW} and underpins a wide range of applications in autonomous driving~\cite{drive1,drive2,uav}, embodied artificial intelligence~\cite{robot1,robot2}, and human–computer interaction~\cite{human1,human2}. Despite the significant progress of RGB-only trackers, they can still drift or miss the target in complex scenarios due to the inherent limitations of visible-spectrum sensors, such as background clutter and low illumination. To enhance robustness, recent works introduce auxiliary modalities and develop multi-modal tracking tasks, including RGB-Thermal (RGB-T)~\cite{lasher,rgbtbenchmark,li2020challenge}, RGB-Depth (RGB-D)~\cite{rgbd1k,DepthTrack,ying2024temporal}, and RGB-Event (RGB-E)~\cite{visevent,swineft}.
\begin{figure}[htbp]
\centering
  \includegraphics[width=0.8\textwidth]{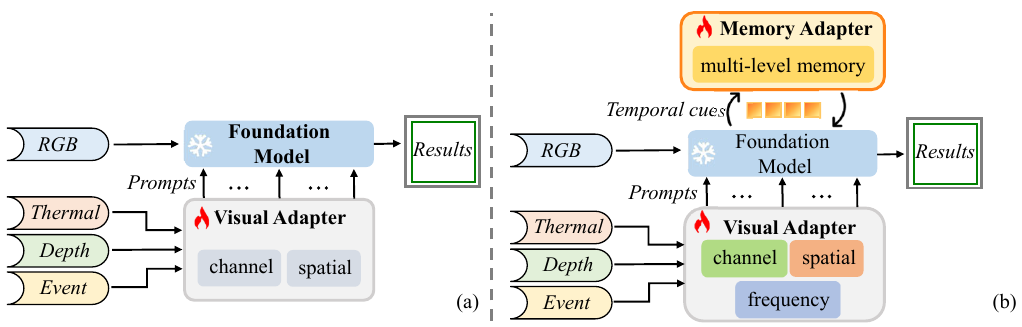}   
  \caption{Framework comparisons between the existing prompt-learning-based tracker and our tracker. (a) Existing trackers propagate temporal cues from adjacent frames and fuse multi-modal features in channel and spatial dimensions. (b) The proposed method integrates a memory adapter to propagate cues adaptively and merge features in channel, spatial, and frequency dimensions.}   
  \label{fig:intro}  
        % \vspace{-15pt}
\end{figure}
Multi-modal tracking methods generally fall into two paradigms. The first is the classic dual-branch architecture~\cite{mtnet,tbsi,DepthTrack}, typically tailored to a specific multi-modal setting and trained end-to-end with full fine-tuning. While effective at capturing modality-specific representations, such methods are constrained by limited training data and high computational cost. The second paradigm adopts prompt learning~\cite{sdstrack,VIPT,onetracker}, which fine-tunes lightweight adapters on frozen RGB-based backbones, improving adaptability and training efficiency. Despite notable progress, two critical gaps remain. (1) Frequency cues are essential in multi-modal tracking~\cite{frequencyselect,frequency_sod}, yet current prompt-learning trackers underuse frequency-domain information and thus struggle to bridge the modality gap. Different modalities exhibit complementary frequency characteristics due to their sensing principles (Fig.~\ref{fig:intro2}): RGB encodes fine-grained high-frequency textures; Thermal and Depth emphasize low-frequency structural contours; Event streams carry sparse, sharp high-frequency motion edges. Direct spatial-domain fusion is therefore suboptimal owing to cross-modality noise and misaligned frequency content. Although recent efforts explore frequency-aware designs~\cite{frequency1,frequency2}, they largely rely on hand-crafted frequency separation, limiting data-driven exploitation. (2) Temporal cues are equally crucial, yet existing trackers~\cite{tctrack,TPAMIYANG,ASTMT} typically restrict modeling to adjacent-frame propagation or confidence-based template updates, which fail to capture long-range dependencies; errors accumulate under occlusion, fast motion, or appearance changes, degrading robustness.

% To address these two core challenges, we propose a unified multi-modal tracker based on a dual adapter architecture, comprising a visual adapter and a memory adapter, as illustrated in Figure~\ref{fig:intro} (b). 
% To facilitate effective modal fusion, we propose a frequency selector that adaptively enhances frequency components for intra-modal feature refinement. In addition, a multi-modal fusion module is designed to generate modality-aware prompts through joint spatial and channel attention, allowing efficient and robust cross-modal interaction.

% To model temporal dependencies, we propose the memory adapter inspired by the human memory mechanism~\cite{Xmem,memory}. We design a multi-level memory pool to store the temporal cues, consisting of short-term, long-term, and permanent memory, which facilitates the object searching process. We also introduce the memory update and retrieval operations. The update operation refreshes the memory pool with each frame to ensure reliability, while the retrieval operation retrieves the appropriate temporal cues for subsequent tracking. 
% Extensive experiments on mainstream multi-modal tracking benchmarks demonstrate that our tracker outperforms both fully fine-tuned and existing prompt-learning-based methods in terms of accuracy and robustness. 

To address these two core challenges, we propose a unified multi-modal tracker that learns frequency and memory-aware prompts via a dual-adapter architecture comprising a frequency-guided visual adapter and a multi-level memory adapter (Fig.~\ref{fig:intro}(b)). 
For effective cross-modal fusion, the visual adapter introduces a frequency selector that adaptively emphasizes informative subbands to refine intra-modal features, and a multi-modal fusion module that produces modality-aware prompts through joint spatial and channel attention, thereby narrowing the modality gap without full fine-tuning.

To model temporal dependencies, the memory adapter is inspired by the human memory mechanism~\cite{Xmem,memory}. It maintains a multi-level memory pool (short-term, long-term, and permanent) to store global temporal cues, together with update and retrieval operations that refresh the memory per frame and select reliable cues for subsequent tracking. This design learns memory-aware prompts that propagate consistent temporal context across frames and recover from occlusion, fast motion, and illumination changes. 
Extensive experiments on mainstream multi-modal tracking benchmarks demonstrate that our tracker outperforms fully fine-tuned and prompt-learning baselines in both accuracy and robustness, while preserving the efficiency of adapter-based tuning.

In summary, our main contributions are as follows:

$\bullet$ We present a unified dual-adapter framework that learns frequency and memory-aware prompts for multi-modal tracking, showing consistent gains on RGB-T, RGB-D, and RGB-E tasks.

$\bullet$ We develop a lightweight frequency-guided visual adapter that aggregates informative cues across frequency, spatial, and channel dimensions to produce modality-aware prompts and enhance cross-modal fusion.

$\bullet$ We propose  a multi-level memory adapter that stores and retrieves global temporal cues with update and retrieval operations, enabling adaptive propagation of temporal context along video sequences.

\section{Related Work}
\subsection{Multi-modal Object Tracking}

Multi-modal object tracking incorporates additional modality with the RGB modality, such as thermal, depth, or event data, to enhance the perceptual capabilities of visible sensors, particularly in complex scenarios where RGB modality may be invalid~\cite{rgbtREVIEW,arkittrack,MMTRACK,rgbt1,PTM,amnet,hu2025exploiting}. 

Most multi-modal trackers are typically extensions of strong RGB-based trackers, encompassing multi-modal feature extraction and fusion to strengthen representations. Previous works follow the two-parallel branch architecture. For example, 
APFNet~\cite{apfnet} explored fusion strategies under various challenging attributes to boost tracking accuracy. 
MTNet~\cite{mtnet} leveraged Transformers to establish the global association for multi-modal feature interaction and reinforcement. Likewise, SPT~\cite{rgbd1k} applied Transformers both in feature extraction and fusion, maximizing the utilization of complementary multi-modal data. However, these methods relying on parallel feature extraction structures for both modalities can introduce considerable computational overhead and training complexity, complicating cross-modal transfer processes.
More recently, prompt learning-based methods have emerged, aiming to develop lightweight adapters on the powerful foundation models to fine-tune them, thereby reducing training costs and enhancing scalability~\cite{VIPT,protrack,untrack}. 
For instance, ViPT~\cite{VIPT} proposed an adapter based on prompt learning, enabling efficient multi-modal fusion in tracking tasks. 
SDSTrack~\cite{sdstrack} introduced a data augmentation for lower-quality modalities, improving tracking performance on specific challenging attributes. 
OneTracker~\cite{onetracker} further expanded the input of multi-modal trackers by incorporating text prompts. SUTrack~\cite{sutrack} designs a unified tracking framework capable of handling a wide range of single-modal and multi-modal tracking tasks.

Despite the remarkable achievements, existing methods often ignore the importance of frequency information, limiting their performance. This motivates us to propose a novel visual adapter that fully extracts multi-modal cues from multiple dimensions, with a particular emphasis on selecting high- and low-frequency characteristics.
% Prompt-learning-based methods effectively leverage the strength of RGB trackers, broadening their applicability to multi-modal tracking.

% We proposed a novel prompt learning-based method which fully explored multi-modal cues from frequency, spatial and channel dimensions.
\subsection{Temporal Modeling in Visual Tracking}
It is well recognized that rich temporal information is essential for visual object tracking, and effectively enhancing context propagation in the temporal domain remains a central focus of current research.

One type of method focuses on designing template update strategies to replace the initial target template over time. 
For instance, Stark~\cite{stark} combined the initial template features with online information to achieve an adaptive template update. 
Yang et al.~\cite{TPAMIYANG} introduced a multi-frame template pool to select the optimal template, mitigating the unreliability of single-frame templates. 
SDSTrack~\cite{sdstrack} applied the template updated mechanism in the tracking framework by using confidence scores to choose appropriate templates. While these methods improve tracking robustness to some extent, they still treat tracking as a frame-by-frame template-matching task without leveraging deeper temporal correlations. 
Another type of method~\cite{seqtrack,seqtrackv2,mamemory,tra3,tra4} attempts to propagate temporal context across frames. 
For instance, TCTrack~\cite{tctrack} propagated template cues between adjacent frames to guide more precise template feature extraction in subsequent frames. 
ODTrack~\cite{odtrack} incorporated global tokens into the attention mechanism, improving temporal propagation efficiency. 
ASTMT~\cite{ASTMT} applied a propagation network in infrared tracking, enhancing temporal transmission. 
SeqTrack~\cite{seqtrackv2}, by contrast, feeds the entire sequence into the tracker and performs tracking from a global perspective.

While these advanced methods underscore the value of continuous temporal information in tracking, relying solely on adjacent-frame propagation risks being misled by noisy or erroneous data.
Unlike existing trackers, our proposed memory adapter investigates global tracking cues, adaptively propagating the temporal relationships among successive frames for more robust tracking. 

\begin{figure*}[t]
  \centering
  \includegraphics[width=0.95\textwidth]{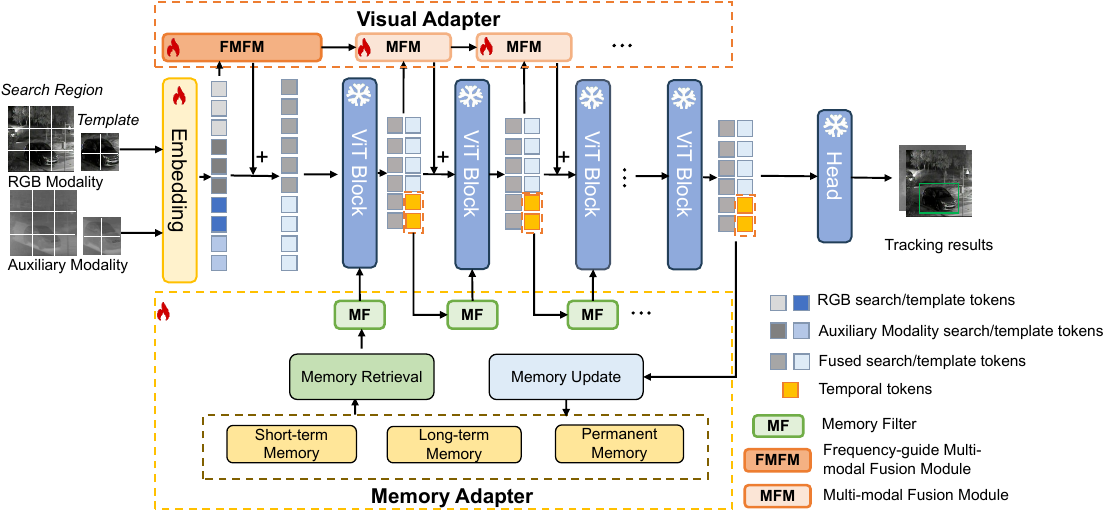}   
  \caption{The framework of the proposed method. We first transform the templates and search region of each modality into tokens, then concatenate them with temporal cue tokens and feed them into the $L$-layer ViT block. The visual adapter and memory adapter are paralleled with the ViT block. The memory adapter is used to propagate the valuable temporal cues across frames, and the visual adapter is used for modality interaction and fusion. The output features are fed into the prediction head to produce the tracking results.}   
  \label{fig:pipeline}  
\end{figure*}

\section{Method}
\subsection{Preliminary and Notation}
\textbf{Problem Formulation.} Given a pair of multi-modal sequence and an initial bounding box, the multi-modal tracking task can learn a tracker $T: \{\textbf{X}_{rgb}^t, \textbf{X}_x^t, \textbf{Z}_{rgb}^0, \textbf{Z}_x^0\} \rightarrow \textbf{B}^t$, where $\textbf{X}_{rgb}^t, \textbf{X}_x^t$ represent the $t$-th search frames of RGB and auxiliary modality (\emph{e.g.,} thermal, depth, or event), $\textbf{Z}_{rgb}^0, \textbf{Z}_x^0$ are the template of RGB and auxiliary modality generated by the initial bounding box $\textbf{B}^0$, $\textbf{B}^t$ represents the $t$-th predicted bounding box.

To fully harness the potential of prompt learning, we propose a visual adapter $V$ and integrate the multi-template mechanism~\cite{odtrack} into multi-modal tracking.  Additionally, to effectively capture and propagate temporal tracking cues, we propose a memory adapter $M$. The overall tracking process can be described as follows:
\begin{equation}
    \begin{aligned}
    M: &\textbf{C}^{t-1} \rightarrow \mathcal{U}\{M\}, \\
    &\mathcal{R}\{M\}\rightarrow \textbf{C}^t,
    \end{aligned}
  \end{equation}
\begin{equation}
    \begin{aligned}
    T:\{ V\{\textbf{X}_{rgb}^t, \textbf{X}_x^t\},V\{\textbf{Z}_{rgb}^0, \textbf{Z}_x^0, ..., \textbf{Z}_{rgb}^i, \textbf{Z}_x^i\}, \textbf{C}^t \}
    \rightarrow \textbf{B}^t,
    \end{aligned}
  \end{equation}
where $\textbf{C}^t$ is the $t$-th temporal tracking cue, $\mathcal{U}\{\cdot\}$ and $\mathcal{R}\{\cdot\}$ represent memory update and retrieval operation, respectively. $\textbf{Z}_{rgb}^t$ and $\textbf{Z}_x^t$ denote the $t$-th templates generated by the corresponding frames and tracking result.

\textbf{Foundation Model.} We choose a powerful RGB tracker ODTrack~\cite{odtrack} as the foundation model. Given the input search region and template $\textbf{X}_{rgb}$ and $\textbf{Z}_{rgb}$, they first sent into patch embedding layer to obtain 1D tokens $\mathcal{H}_{x}^0$ and $\mathcal{H}_{z}^0$. These tokens are then concatenated to form the input token $\mathcal{H}^0$ = [$\mathcal{H}_{x}^0$, $\mathcal{H}_{z}^0$]. The input tokens are fed into $L$-layers vision transformer block encoder and the output is passed through a box head to generate the tracking results. The prorogation process can be formulated as follows:

 \begin{equation}
    \begin{aligned}
    \mathcal{H}^l &= \mathcal{E}^l(\mathcal{H}^{l-1}), l=1,2,3,...,L,\\
    \textbf{B} &= \phi(\mathcal{H}^L),\\
    \end{aligned}
  \end{equation}
where $\mathcal{E}(\cdot)$ is the vision transformer block and $\phi(\cdot)$ is prediction head.

\subsection{Overall Framework}
The proposed framework is illustrated in Figure~\ref{fig:pipeline}. It consists of four main components: the ViT backbone, visual adapter, memory adapter, and prediction head. Initially, the templates and search regions for both RGB and auxiliary modalities are embedded into tokens through the patch-embedding layer. These tokens are then sent to the frequency-guide multi-modal fusion module (FMFM) for shallow feature fusion. Subsequently, the temporal tracking cue tokens are retrieved from the multi-level memory pool, passed through a memory filter, and then sent to the ViT block along with the search region and template tokens. After each ViT block, the output undergoes the multi-modal fusion module (MFM) for modality enhancement and fusion, while the temporal tracking cues pass through the memory filter. After passing through $L$ layers of ViT blocks, the final tokens are used in the head operation to obtain the tracking results, and temporal tracking cues are stored in the multi-level memory pool.
\begin{figure}[t]
\centering
  \includegraphics[width=0.67\textwidth]{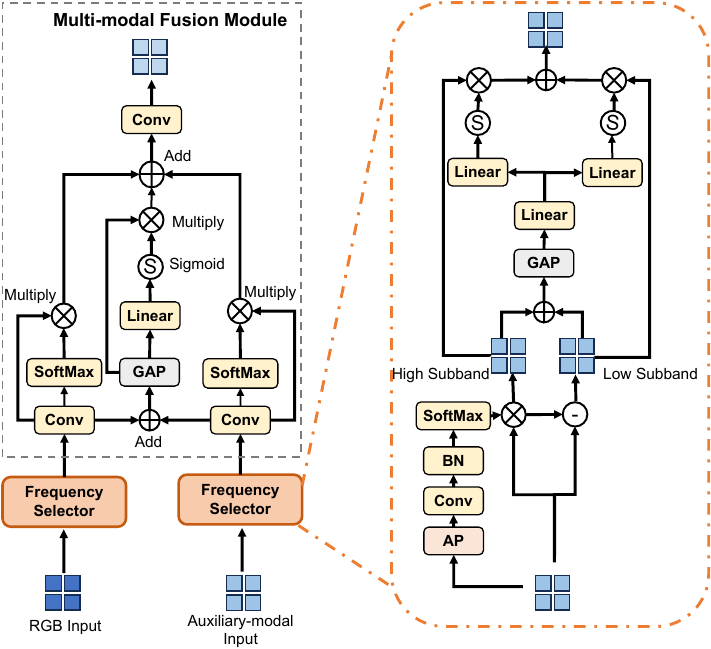}   
  \caption{Detailed design of the frequency-guide multi-modal fusion module, which enhances the feature representation by combining spatial, channel, and frequency information from different modalities.} 
  \label{fig:adapter}  
\end{figure}

\subsection{Visual Adapter}
The visual adapter plays a crucial role in prompt-learning-based multi-modal tracking methods, as it directly influences how effectively multi-modal information is leveraged. Our objective is to design a visual adapter that fully explores the potential of multi-modal data while maintaining efficiency. To achieve this, we propose a frequency-guided multi-modal fusion module for the first layer of the visual adapter, followed by multi-modal fusion modules for subsequent layers. With this design, we can extract frequency information in the shallow layer while adhering to the prompt-learning principle of maintaining parameter efficiency.

\textbf{Frequency Selector.} Frequency is a key attribute in images, where the high-frequency subbands typically contain edge or contour information, while the low-frequency subbands generally include detailed features. The frequency selector is designed to extract rich texture details from RGB data and edge or contour information from auxiliary modalities. By selectively enhancing these frequency components, the frequency selector facilitates better integration of multi-modal features, improving tracking accuracy across varying conditions.
% To leverage this, we designed a frequency selector to distinguish and enhance the frequency components of the image. 
The detailed design is shown in the right part of Figure~\ref{fig:adapter}, inspired by other frequency decomposition methods~\cite{frequencyselect,frequency_sod}, we first separate the input into high-frequency and low-frequency components, which can be calculated as follows:
\begin{align}
\textbf{F}_{high} &= \textbf{F}_{ori} \otimes (\mathrm{Softmax}(\mathrm{BN}(\mathrm{Conv}(\mathrm{Ap}(\textbf{F}_{ori}))))), \label{eq1} \\
\textbf{F}_{low} &=\textbf{F}_{ori} - \textbf{F}_{high}, \label{eq3} 
\end{align}
where $\mathrm{Ap}(\cdot)$, $\mathrm{Conv}(\cdot)$, and $\mathrm{BN}(\cdot)$ represent average pooling, convolution and batch normalization, respectively. $\textbf{F}_{ori}$, $\textbf{F}_{high}$, and $\textbf{F}_{low}$ denote the input feature, the high- and low-frequency features, respectively. Then, we select and fuse the different frequency features to get the more representative features, which can be calculated as follows:
\begin{align}
\textbf{F}_{global} &= \mathrm{FC}\left(\mathrm{GAP}\left( \textbf{F}_{high} \oplus \textbf{F}_{low} \right)\right), \label{eq1} \\
\hat{\textbf{F}_{high}} &= \sigma\left(\mathrm{FC}_{high}\left(\textbf{F}_{global}\right)\right) \otimes \textbf{F}_{high}, \label{eq3} \\
\hat{\textbf{F}_{low}} &= \sigma\left(\mathrm{FC}_{low}\left(\textbf{F}_{global}\right)\right) \otimes \textbf{F}_{low}, \label{eq4} 
\end{align}
where $\mathrm{FC}(\cdot)$ is the linear layer, $\mathrm{GAP}(\cdot)$ is global average pooling, $\sigma$ is Sigmoid, $\oplus$ denotes element-wise addition and $\otimes$ denotes the element-wise multiplication operation. Finally, we use element-wise addition to combine the high-frequency and low-frequency components of the image.

\textbf{Multi-modal Fusion Module.} The multi-modal fusion module integrates the multi-modal information from both spatial and channel perspectives. As shown in the left part of Figure~\ref{fig:adapter}, the input is divided into three branches: two branches are dedicated to enhancing multi-modal features from a spatial perspective, highlighting the most informative features, while the third branch concatenates the dual modalities and selects the most relevant channels, effectively suppressing the influence of redundant information, which can be calculated as follows:
\begin{align}
\textbf{F}_i & = \mathrm{Conv}(\textbf{I}_{i}), \quad i \in \{RGB, X\},\label{eq1}\\
\textbf{F}^{s}_i &= \textbf{F}_i\otimes(\mathrm{Softmax}(\textbf{F}_{i}),\quad i \in \{RGB, X\}, \label{eq1} \\
\textbf{F}^{g} &= \mathrm{GAP}( \textbf{F}_{RGB} \oplus \textbf{F}_{X}), \label{eq2} \\
\textbf{F}^{c} &= \textbf{F}^{g}\otimes(\sigma(\mathrm{FC}(\textbf{F}^{g})),\label{eq3} 
\end{align}
where $\textbf{F}^s$, $\textbf{F}^c$ represent the features passed through the spatial and channel branches, respectively. $\textbf{I}_{RGB}$ and $\textbf{I}_{X}$ are the inputs of RGB and auxiliary modality from the frequency selector in the first layer or those from the ViT backbone in subsequent layers. We then apply element-wise addition to combine the outputs of the three branches, followed by a convolution layer to produce the final output. This output is added to the original output from the previous ViT block and concatenated with the temporal tracking cue token, which serves as the input for the next ViT block.

We thoroughly exploit the potential of modality fusion from frequency, spatial, and channel perspectives, allowing us to fine-tune only a small number of parameters for prompt learning. This approach achieves impressive performance across a wide range of multi-modal tracking tasks.

\begin{figure}[t]
\centering
  \includegraphics[width=0.78\textwidth]{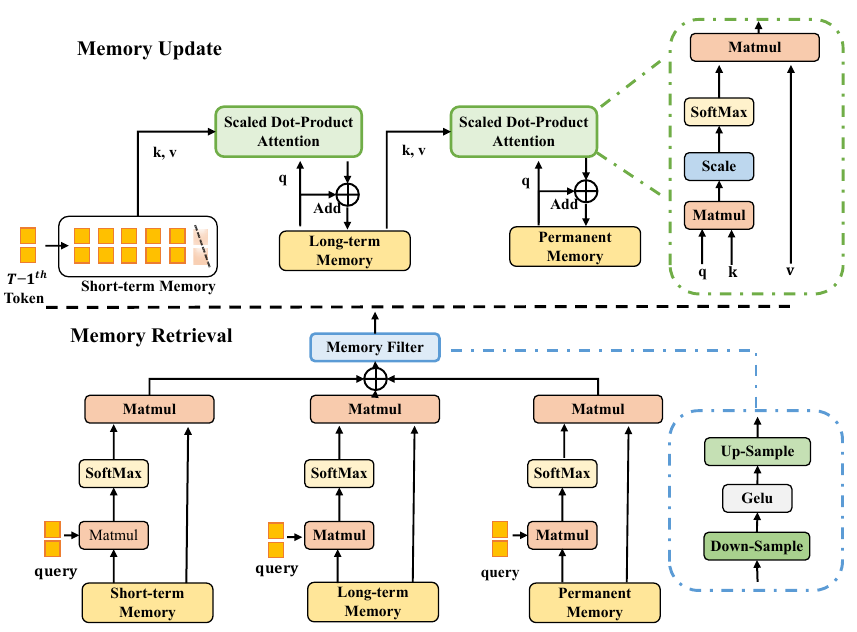}   
  \caption{Detailed design of memory update and memory retrieval, which ensures the most reliable tracking cues are propagated in the subsequent sequence.}   
  \label{fig:memory}  
\end{figure}
\subsection{Memory Adapter}
Memory mechanisms are commonly used in VOT to store temporal tracking cues~\cite{odtrack,tctrack}. To further improve the robustness of multi-modal object tracking, we propose a memory adapter inspired by the human memory system, consisting of short-term memory, long-term memory, and permanent memory. The proposed multi-level memory operates through two key operations: memory update and memory retrieval. In the memory update operation, we use the previous tracking cue token to update all memory levels, while in the memory retrieval operation, the $T$-th temporal tracking cue is retrieved from the memory. Each level holds an \(N \!\times\! H\) tensor, where \(H\!=\!768\) aligns with the ViT token dimension and \(N\) is set to 8, 8, and 3 for short-term, long-term, and permanent memory, respectively.  
The short-term memory stores tokens from the most recent eight frames, while the long-term and permanent memory banks are hierarchically refreshed following a temporal update protocol.

% Short-term memory updates rapidly, storing recent information; long-term memory retains information from more distant past events; and permanent memory stores high-quality, selected memories.

The memory update and memory retrieval operations are shown in Figure~\ref{fig:memory}. In the memory update operation, the initial cue tokens are stored in each level of memory at the start of tracking. After each frame during the tracking process, the temporal tracking cue token is first used to update the short-term memory, which stores the most recent 8 frames of temporal tracking cue tokens. Next, a scaled dot-product attention~\cite{attention} is applied to extract high-quality memory from the short-term memory to update the long-term memory, which can be calculated as follows:
\begin{align}
\textbf{LTM}^{'} &= \mathrm{Softmax}\left(\frac{\textbf{Q}\cdot \textbf{K}^{T}}{\sqrt{d_{k}}}\right) \cdot \textbf{V},\\
\textbf{LTM} &= \textbf{LTM} \oplus \textbf{LTM}^{'},
\end{align}
where $\textbf{Q}$, $\textbf{K}$, and $\textbf{V}$ represent the query, key, and value, respectively, the key and value are derived from the short-term memory, while the query comes from the long-term memory, $d_{k}$ represents the dimension of the key, the $\oplus$ represents element-wise addition. The update operation for the permanent memory follows the same procedure as that for the long-term memory.

In the memory retrieval operation, we aim to extract global tracking information from the memory while avoiding errors from incorrect memories. To achieve this, we use the latest temporal tracking cue as the query to select information from each level of memory. This selection operation can be calculated as follows:
\begin{align}
\textbf{W}_{i} & = \mathrm{Softmax}(\textbf{q}\cdot \textbf{M}_i^{T}), \quad i \in \{s, l, p\}, \label{eq2} \\
\textbf{C}_{i} & = \textbf{W}_i \otimes \textbf{M}_i, \quad i \in \{s, l, p\}, \label{eq3} 
\end{align}
where $\mathrm{Softmax}(\cdot)$ denotes the Softmax operation, \textbf{q} denotes the query. $\textbf{W}_{s}$, $\textbf{W}_l$, and $\textbf{W}_p$ represent the weights of each memory cue in short-term memory, long-term memory, and permanent memory, respectively. 
Likewise, $\textbf{M}_{s}$, $\textbf{M}_l$, $\textbf{M}_p$ refer to the stored memory in short-term, long-term, and permanent memory, while $\textbf{C}_{s}$, $\textbf{C}_{l}$, and $\textbf{C}_{p}$ represent the selected memory from these levels, respectively. After selecting from each level of memory, we use element-wise addition to combine the results and then send them into the memory filter, which can be calculated as follows:
\begin{align}
\textbf{C} &= \textbf{C}_s \oplus \textbf{C}_l \oplus \textbf{C}_p, \label{eq1} \\
\textbf{C}^{'} &=\mathrm{Us}(\mathrm{G}(\mathrm{Ds}(\textbf{C}))), \label{eq3} 
\end{align}
where $\mathrm{Ds}(\cdot)$ and $\mathrm{Us}(\cdot)$ represent the downsampling and upsampling operations, respectively. $\mathrm{G}(\cdot)$ is the GELU activation function. The memory filter is applied after each ViT block to ensure that the temporal tracking cue is appropriately maintained and adjusted for each level of the ViT block.

\subsection{Prediction Head and Loss Function}
We adopt the prediction head from the base tracker~\cite{odtrack}, freezing both the parameters of classification and regression heads. The classification head yields the score map and the regression head generates the bounding box. These components work together to achieve the final tracking outcome.

The loss function contains classification loss and regression loss.
The proposed method employs focal loss~\cite{rgbtbenchmark} as the classification loss $\mathcal{L}_{cls}$, which is suitable for the dataset with long-tail distribution and can be calculated as:

\begin{equation}
\begin{aligned}
\mathcal{L}_{cls} = -\sum_{t} \alpha \left ( 1-p_{t} \right ) ^{\gamma } log(p_{t}),
\end{aligned}
\end{equation}
where $t$ represents the $t$-th samples, $\alpha_t$ is the weight coefficient, $p_{t}$ indicates the probability belonging to the foreground.
The regression loss contains $\mathcal{L}_{1}$ loss and $GIoU$~\cite{GIOU} loss, which can be calculated as:
\begin{equation}
\begin{aligned}
\mathcal{L}_{reg} = \sum_{t} \left ( \lambda_1 \mathcal{L}_1\left ( b_t,\hat{b_t}  \right )  + \lambda_2 \mathcal{L}_{GIoU}\left ( b_t,\hat{b_t}  \right )  \right ) ,
\end{aligned}
\end{equation}
where $\lambda_1$ and $\lambda_2$ are regularization parameters, which are set as 5 and 2, respectively.  $b_t$ denotes the $t$-th predicted bounding box, $\hat{b_t}$ is the corresponding ground truth.
The overall loss can be expressed as follows:
\begin{equation}
\begin{aligned}
\mathcal{L} = \mathcal{L}_{cls} + \mathcal{L}_{reg}.
\end{aligned}
\end{equation}

\section{Experiment}

\subsection{Datasets and Metrics}
To fully validate the effectiveness of our method, we evaluate it on existing representative multi-modal tracking tasks, including RGB-T tracking, RGB-D tracking and RGB-E tracking. We use the standard evaluation metrics for each task to validate the performance of the proposed method. 

\textbf{RGB-T tracking.} For RGB-T tracking, we evaluate our method on the latest RGB-T tracking datasets, RGBT234~\cite{rgbtbenchmark} and LasHeR~\cite{lasher}, which are the largest RGB-T tracking datasets with more than 200 testing video sequences of different challenging attributes. We use Precision Rate (PR) and Success Rate (SR) as primary measures and the threshold of center location error is set to 20 pixels. 
PR represents the ratio of frames $f_p$ with center error smaller than a threshold to the total number of frames $N$, which can be calculated as:
\begin{equation}
PR = \frac{N_p}{N},
\end{equation}
where $N_p$ represents the number of frames with center error smaller than a threshold; $N$ represents the total number of the sequence.
SR is defined as the ratio of frames $s_p$ with IoU exceeding a certain threshold to the total number of frames $N$, and it can be calculated as:
\begin{equation}
SR = \frac{N_s}{N},
\end{equation}
where $N_s$ represents the number of frames with IoU exceeding a certain threshold.

\textbf{RGB-D tracking.} For RGB-D tracking, we conduct experiments on the DepthTrack~\cite{DepthTrack} and VOT22-RGBD~\cite{vot23} datasets. DepthTrack is a large-scale, long-term RGB-D tracking dataset consisting of 200 pairs of RGB-D videos and the VOT-RGBD22 dataset is the latest RGB-D tracking dataset proposed in the VOT challenge~\cite{vot23}, containing more than 140 test sequences. For evaluation on DepthTrack, we use Precision (Pre), Recall (Re) and F-score as metrics, while for VOT22-RGBD, we adopt Accuracy (A), Robustness (R) and Expected Average Overlap (EAO) to assess performance. 
Precision is calculated by the Gaussian Mixture Distribution between all frame output boxes and the given correct output boxes. The sum of all computed Gaussian Mixture Distributions is divided by the total frame count to determine tracking precision. The precision is calculated as follows:
\begin{equation}
\label{rgequ:1}
\operatorname{Pre}\left(\tau_\theta\right)=\frac{1}{N_p} \sum_{t} \Omega\left(A_t\left(\theta_t\right), G_t\right), t \in\left\{t: A_t\left(\theta_t\right) \neq \emptyset\right\} ,
\end{equation}
where $\operatorname{Pre}\left(\tau_\theta\right)$ 
represents precision, $A_t\left(\theta_t\right)$ represents the tracker's output, $G_t$ represents the ground truth, and 
$\Omega(\cdot)$ represents the intersection of the two. The sum is taken over all non-empty predicted results.

Recall is calculated by the Gaussian Mixture Distribution between all frame output boxes and the given correct output boxes. The sum of all computed Gaussian Mixture Distributions is divided by the total frame count where targets are present to determine tracking recall:
\begin{equation}
\label{rgequ:2}
\operatorname{Re}\left(\tau_\theta\right)=\frac{1}{N_g} \sum_{t} \Omega\left(A_t\left(\theta_t\right), G_t\right), t \in\left\{t: G_t \neq \emptyset\right\} ,
\end{equation}
where $\operatorname{Re}\left(\tau_\theta\right)$ represents recall, $A_t\left(\theta_t\right)$ represents the tracker's output. The sum is taken over all non-empty ground truth results.

F-score is divided by the summary of Pr and Re and then multiplied by two to obtain the tracking F-score:
\begin{equation}
\label{rgequ:3}
F\text{-}score\left(\tau_\theta\right)=2\frac{\operatorname{Pr}\left(\tau_\theta\right) \operatorname{Re}\left(\tau_\theta\right)}{\left(\operatorname{Pr}\left(\tau_\theta\right)+\operatorname{Re}\left(\tau_\theta\right)\right)} ,
\end{equation}
where $\operatorname{Re}\left(\tau_\theta\right)$  represents the corresponding recall;  $\operatorname{Pre}\left(\tau_\theta\right)$ represents the corresponding precision.

\textbf{RGB-E tracking.} For RGB-E tracking, we report the experimental results on VisEvent~\cite{visevent}, which is the largest RGB-E tracking dataset, containing over 500 sequences. The metrics that we use to evaluate are Precision Rate (PR) and Success Rate (SR), just the same as RGB-T tracking. 

\subsection{Experimental Settings}
When fine-tuning the proposed method, we choose the training sets of LasHeR for RGB-T tracking, DepthTrack for RGB-D tracking, and VisEvent for RGB-E tracking. The proposed method is trained on one NVIDIA RTX 4090 GPU with a batch size of 16. We use the ViT-base trained on MAE as 
our baseline. The training consists of two stages: in the first stage, we fine-tune the visual adapter and patch-embedding layer for 60 epochs while freezing the other part of the network. In the second stage, we fine-tune the memory adapter on top of stage one for another 60 epochs and freeze the other part except for the patch-embedding layer and visual adapter. Each epoch contains 10,000 samples and we use the AdamW optimizer with a learning rate of $5e^{-4}$. In addition, our method contains a total of 98.9M parameters, introducing 7.3M more than the baseline, with an additional computational cost of 1 GFLOPS.

\subsection{Comparison with the State-of-the-Art Methods}
We compare the proposed method with two categories of state-of-the-art multi-modal trackers: traditional two-branch methods and prompt-learning methods. The former includes methods specifically designed for a particular \begin{table}
    \centering
    \setlength{\tabcolsep}{1.5pt}
    \small
    \caption{Comparison between the proposed method and the state-of-the-art trackers on RGB-T datasets. The best results are highlighted in \textbf{bold}. The performance is evaluated in terms of Precision Rate (PR) and Success Rate (SR).}
    \label{tablergbt}
    \resizebox{0.8\linewidth}{!}{
    \begin{tabular}{c|r|c|cc|cc|c|c}
        \toprule
        &\multirow{2}{*}{Methods} &\multirow{2}{*}{Publication}& \multicolumn{2}{c|}{RGBT234} & \multicolumn{2}{c|}{LasHeR} &\multirow{2}{*}{Param} &\multirow{2}{*}{FPS}\\
       & & & PR & SR  & PR & SR &&\\
        \midrule
      \multirow{6}{*}{\rotatebox{90}{Traditional}}  & TBSI~\cite{tbsi}&$\text{CVPR23}$ & 0.871 & 0.637 &  0.692 & 0.556 & 350 &36   \\
        &MTNet~\cite{mtnet}&$\text{ICME23}$ & 0.850 & 0.619 & 0.608 & 0.474 &-&55\\
         &GMMT~\cite{gmmt}&$\text{AAAI24}$ & 0.879 & 0.647 & 0.707 & 0.566 & - & - \\
        &STMT~\cite{STMT}&$\text{TCSVT24}$ & 0.865 & 0.638 & 0.674 & 0.537 & - &39  \\
        &CAT++~\cite{cat++}& $\text{TIP24}$ & 0.840 & 0.592 &  0.509 & 0.356 & 90 &14\\ 
        &CAFormer~\cite{caformer}& $\text{AAAI25}$ & 0.883 & 0.664 &  0.700 & 0.556 & - &84\\\midrule
        \multirow{7}{*}{\rotatebox{90}{Prompt} }&ProTrack~\cite{protrack}&$\text{MM23}$& 0.795 & 0.599 & 0.538 & 0.420 & - &30\\
        &ViPT~\cite{VIPT}&$\text{CVPR23}$& 0.835 & 0.617 & 0.651 & 0.525 & 93 & 25  \\
        &SDSTrack~\cite{sdstrack}&$\text{CVPR24}$ & 0.848 & 0.625 & 0.665 & 0.531 & 107.8 & 21 \\
        &UN-Track~\cite{untrack}&$\text{CVPR24}$ & 0.837 & 0.618 & 0.667 & 0.536 & 92.1 & -\\
        &OneTracker~\cite{onetracker}&$\text{CVPR24}$ & 0.857 & 0.642 & 0.672 & 0.538 & 99.8 & - \\
        &TaTrack~\cite{tatrack}&$\text{AAAI24}$ & 0.872 & 0.644 & 0.702 & 0.561 & - & 26\\
        &BaT~\cite{bat}&$\text{AAAI24}$ & 0.868 & 0.641 & 0.702 & 0.563  &- &-\\ 
        &IPL~\cite{IPL}& $\text{IJCV25}$ & 0.883 & 0.657 &  0.694 & 0.553 & - &-\\
        &CMDTrack~\cite{CMDTrack}& $\text{TPAMI25}$ & 0.859 & 0.618 &  0.688 & 0.566 & - &67\\\midrule
         
        &\textbf{Ours}& - & \textbf{0.919} & \textbf{0.689} & \textbf{0.726} & \textbf{0.571} & 98.9 & 65 \\ 
        \bottomrule
\end{tabular}}
\end{table}
type of multi-modal tracking that fully fine-tunes the two-branch network for both modalities. The latter encompasses methods designed for general multi-modal tracking that only fine-tune the adapters and patch-embedding layers.

\textbf{RGB-T tracking.} The comparison results for RGB-T tracking are shown in Table~\ref{tablergbt}, where the proposed method outperforms all existing methods in terms of precision rate and success rate. Specifically, it exceeds the second-best method, TaTrack~\cite{tatrack}, on the RGBT234 dataset by 4.7\% and 4.5\% gains in PR and SR, respectively. On the LasHeR dataset, the proposed method achieves improvements of 2.4\% and 0.8\% in PR and SR compared to BaT~\cite{bat}. Additionally, compared to the latest two-branch tracker, STMT~\cite{STMT}, the proposed method shows enhancements of 5.2\% and 3.4\% in PR and SR, respectively. 
\begin{table}
    \centering
    \small
    \setlength{\tabcolsep}{1.8pt}
    \caption{Comparison between the proposed method and the state-of-the-art trackers on RGB-D datasets. The best results are highlighted in \textbf{bold}. The performance is evaluated in terms of precision (Pre), recall (Re), F-score(F) on DepthTrack and EAO, accuracy(A), robustness(R) on VOT-RGBD22.}
    \label{tablergbd}
    \resizebox{0.85\linewidth}{!}{
    \begin{tabular}{c|r|c|ccc|ccc}
        \toprule
        &\multirow{2}{*}{Methods} & \multirow{2}{*}{Publication} & \multicolumn{3}{c|}{DepthTrack} & \multicolumn{3}{c}{VOT-RGBD22} \\
        && & Pre & Re  & F & EAO & A  & R\\
        \midrule
        \multirow{5}{*}{\rotatebox{90}{Traditional}}   
        &DeT~\cite{DepthTrack}& $\text{ICCV21}$     & 0.506 & 0.560 &  0.532 & 0.657 &0.760 &0.845  \\
         &SBT-D~\cite{vot23}& $\text{ECCV22}$ & - & - & - & 0.708 &0.809&0.864 \\
         &OSTrack~\cite{ostrack}& $\text{ECCV22}$ & 0.536 & 0.522 &  0.529 & 0.676&0.803 &0.833  \\
         &SPT~\cite{rgbd1k}& $\text{AAAI23}$ & 0.549 & 0.527 &  0.538 & 0.651&0.798 &0.851  \\
         &ARKitTrack~\cite{arkittrack}& $\text{CVPR23}$ & 0.617 & 0.607 &  0.612 & - & - & -  \\
        &TABBTrack~\cite{tabb}& $\text{PR25}$ & 0.622 & 0.615 &  0.618 & - & - & -  \\
               \midrule
       \multirow{5}{*}{\rotatebox{90}{Prompt}}&ProTrack~\cite{protrack}& $\text{MM23}$ &0.583 & 0.573 & 0.578 & 0.651 & 0.801 & 0.802 \\
        &ViPT~\cite{VIPT}& $\text{CVPR23}$ &0.596 & 0.594 & 0.592 & 0.721 & 0.815& 0.871\\
        &SDSTrack~\cite{sdstrack} & $\text{CVPR24}$ & 0.619 &  0.609 &  0.614 & 0.728 & 0.812 & 0.883 \\
        &UN-Track~\cite{untrack} & $\text{CVPR24}$ & 0.560 & 0.557 & 0.558 & 0.721 & 0.815 & 0.871 \\
        &OneTracker~\cite{onetracker} & $\text{CVPR24}$ & 0.609 & 0.604 & 0.607 & 0.721 & 0.819 & 0.872 \\
        &CMDTrack~\cite{CMDTrack} & $\text{TPAMI25}$ & 0.591 & 0.607 & 0.598 & - & - & - \\
         \midrule
        &\textbf{Ours}& - & \textbf{0.636} & \textbf{0.663} & \textbf{0.649} & \textbf{0.773} & \textbf{0.821} & \textbf{0.933} \\ 
        \bottomrule
    \end{tabular}} 
\end{table}
\begin{table}
    \centering
% \normalsize
\small
    \setlength{\tabcolsep}{7pt}
    \caption{Comparison between the proposed method and the state-of-the-art trackers on RGB-E datasets. The best results are highlighted in \textbf{bold}. The performance is evaluated in terms of precision rate (PR) and success rate (SR).}
    \label{tableevent}
    % \resizebox{\linewidth}{!}{
    \begin{tabular}{c|r|c|cc}
        \toprule  
        &\multirow{2}{*}{Methods} &\multirow{2}{*}{Publication}& \multicolumn{2}{c}{VisEvent}  \\
       & & & PR & SR  \\
        \midrule
      \multirow{4}{*}{\rotatebox{90}{Traditional}}   
        &Dimp~\cite{dimp}&$\text{ICCV19}$ & 0.691 & 0.533   \\
        &TansT~\cite{tatrack}&$\text{CVPR21}$ & 0.676 & 0.511   \\
        &OSTrack~\cite{ostrack}&$\text{ECCV22}$ & 0.695 & 0.534 \\
        &SwinEFT~\cite{swineft}&$\text{AI23}$ & 0.710 & 0.565  \\
 \midrule
        \multirow{5}{*}{\rotatebox{90}{Prompt}}&ProTrack~\cite{protrack}&$\text{MM23}$& 0.632 & 0.471 \\
        &ViPT~\cite{VIPT}&$\text{CVPR23}$& 0.758 & 0.592  \\
        &SDSTrack~\cite{sdstrack}&$\text{CVPR24}$ &  0.767 & 0.597   \\
        &UN-Track~\cite{untrack}&$\text{CVPR24}$ & 0.763 & 0.597  \\
        &OneTracker~\cite{onetracker}&$\text{CVPR24}$ &  0.767 &  0.608   \\
        &CMDTrack~\cite{CMDTrack}&$\text{TPAMI25}$ &  0.758 &  0.613   \\\midrule
          &\textbf{Ours}& - & \textbf{0.803} & \textbf{0.626} \\
        \bottomrule
    \end{tabular}
      % \vspace{-15pt}
\end{table}

\textbf{RGB-D tracking.} 
The experimental results for RGB-D tracking are presented in Table~\ref{tablergbd}. The proposed method achieves 63.6\%, 66.3\%, and 64.9\% in precision, recall, and F-score on the DepthTrack dataset, and 77.3\%, 82.1\%, and 93.3\% in EAO, accuracy, and robustness on the VOT-RGBD22 dataset, respectively. These results demonstrate that the proposed tracker significantly outperforms existing methods across all metrics. Notably, it surpasses the second-best method, SDSTrack~\cite{sdstrack}, with gains of 5\% in robustness and 4.5\% in EAO.

\textbf{RGB-E tracking.}
 We compare the proposed method with the state-of-the-art RGB-E trackers. As shown in Table~\ref{tableevent}, the proposed method outperforms the best prompt-learning method, OneTracker~\cite{onetracker} with improvements of 3.6\% in PR and 1.8\% in SR. Additionally, the proposed method achieves gains of 9.3\% in PR and 6.1\% in SR over SwinEFT~\cite{swineft}, which is specifically designed for RGB-E tracking.
These results demonstrate that our tracker also exhibits strong generalization capability in various tracking tasks.

\begin{table}
    \centering
% \normalsize
\small
    \setlength{\tabcolsep}{3pt}
    \caption{Component analysis on multi-modal tracking datasets. Visual represents the proposed visual adapter, and memory represents the memory adapter.}
    \label{tableabla}
    % \resizebox{\linewidth}{!}{
    \begin{tabular}{cc|cc|ccc|cc}
        \toprule
        \multirow{2}{*}{Visual} & \multirow{2}{*}{Memory}& \multicolumn{2}{c|}{LasHeR} &\multicolumn{3}{c|}{DepthTrack} &\multicolumn{2}{c}{VisEvent}\\
       &  & PR & SR & Pre & Re & F & PR & SR\\
        \midrule
&  & 0.659 & 0.518 & 0.586 & 0.608 & 0.598 & 0.784 & 0.605\\
\ding{51} &  & 0.718 & 0.565 & 0.614 & 0.639 & 0.626 & 0.790 & 0.618\\
& \ding{51} & 0.689 & 0.545 & 0.600 & 0.625 & 0.613 & 0.795 & 0.618\\
\ding{51}& \ding{51} & \textbf{0.726} & \textbf{0.571} & \textbf{0.636} & \textbf{0.663} & \textbf{0.649} & \textbf{0.803} & \textbf{0.626}\\
        \bottomrule
    \end{tabular}
\end{table}

% \subsection{Exploration Studies}
% We conduct sufficient experiments on the proposed method, including component analysis, 
% attribute analysis, comparison of tunable parameters,  comparison of visual adapters and visualization analysis, to more comprehensively explore the effectiveness of the proposed method.

\subsection{Ablation Studies.} 
% To assess the contributions of each component in our method, we conduct an ablation study. As shown in Table~\ref{tableabla}, the first row represents the baseline, which directly performs element-wise addition to fuse RGB and the auxiliary modality while fine-tuning the patch-embedding layer. The second row replaces element-wise addition with our proposed visual adapter, showing a significant improvement, particularly in the LasHeR dataset, where PR and SR increased by 5.9\% and 4.7\%, respectively. In the third row, the baseline adds the multi-level memory, resulting in performance gains across all multi-modal tracking datasets and confirming the contribution of the multi-level memory in enhancing tracking robustness. The last row represents the full version of the proposed method, which combined the multi-level memory pool and visual adapter, yielding substantial improvements and demonstrating that these two components work well together to boost tracking performance.
\emph{1) Effectiveness of Components: }
% We conduct an ablation study to evaluate key components of our method. As shown in Table~\ref{tableabla}, the baseline (Row 1) fuses RGB and auxiliary modalities via element-wise addition with fine-tuned patch embedding, yielding initial results. Replacing element-wise addition with our visual adapter (Row 2) significantly improves performance, notably on LasHeR (+5.9\% PR and +4.7\% SR). Integrating the multi-level memory into the baseline (Row 3) enhances robustness across all datasets. The full method (Row 4), combining both components, achieves the highest performance gains, demonstrating their synergistic effectiveness in boosting tracking accuracy.
We conduct an ablation study to evaluate key components of our method. 
As shown in Table~\ref{tableabla}, the baseline (Row~1) fuses RGB and auxiliary modalities via element-wise addition with fine-tuned patch embedding, yielding initial results. 
Replacing element-wise addition with our visual adapter (Row~2) significantly improves performance, notably on LasHeR (+5.9\% PR and +4.7\% SR). On DepthTrack, this change lifts the F-score from 0.598 to 0.626 (+2.8 points), indicating that frequency-aware recalibration benefits sequences where depth maps are noisy but still contain high-frequency edge cues.
Integrating the multi-level memory into the baseline (Row~3) enhances robustness across all datasets. The gain is most evident on VisEvent, where SR rises from 0.605 to 0.618, confirming that temporal aggregation helps smooth the bursty event stream and recover from momentary information loss. 
The full method (Row~4), combining both components, achieves the highest performance gains, demonstrating their synergistic effectiveness in boosting tracking accuracy.
\begin{table}
    \centering
% \normalsize
\small
    \setlength{\tabcolsep}{2pt}
    \caption{Comparison of visual adapters on multi-modal tracking datasets.}
    \label{tableadapter}
    \begin{tabular}{c|c|cc|ccc|cc}
        \toprule
        \multirow{2}{*}{Visual} & \multirow{2}{*}{Params(M)}& \multicolumn{2}{c|}{LasHeR} &\multicolumn{3}{c|}{DepthTrack} &\multicolumn{2}{c}{VisEvent}\\
       &  & PR & SR & Pre & Re & F & PR & SR\\
        \midrule
- & 1.9 & 0.659 & 0.518 & 0.586 & 0.608 & 0.598 & 0.784 & 0.605\\
Fovea & 2.1 & 0.690 & 0.544 & 0.593 & 0.614 & 0.604 & 0.790 & 0.610\\
\textbf{Ours} & 3.8 & \textbf{0.718} & \textbf{0.565} & \textbf{0.600} & \textbf{0.625} & \textbf{0.613} & \textbf{0.795} & \textbf{0.618}\\
        \bottomrule
    \end{tabular} 
\end{table}
\begin{table}
    \centering
% \normalsize
\small
% \scriptsize
    \setlength{\tabcolsep}{7pt}
    \caption{Comparison of tunable parameters on the LasHeR dataset, the performance is evaluated in terms of precision rate (PR) and success rate (SR).}
    \label{tabletun}
    % \resizebox{\linewidth}{!}{
    \begin{tabular}{c|cc|cc}
        \toprule
        \multirow{2}{*}{Method} &Model &Tunable & \multicolumn{2}{c}{LasHeR} \\
       &  Params(M)& Params(M) & PR & SR \\
        \midrule
         Frozen &93.4 &1.9 & 0.659& 0.518 \\
        FFT &93.4 &93.4 & 0.702 & 0.555 \\
        w/o Memory &95.3 &3.8 & 0.718& 0.565 \\
        \textbf{Ours} &98.9 & 7.3 & \textbf{0.726}& \textbf{0.572} \\
        \bottomrule
    \end{tabular}
\end{table}

\emph{2) Effectiveness of Visual Adapter: }
\begin{table}
    \centering
    \small
    \renewcommand{\arraystretch}{0.9} % 调整行间距
    \setlength{\tabcolsep}{4pt}
    \caption{Effectiveness of visual adapter on the LasHeR dataset, the performance is evaluated in terms of precision rate (PR) and success rate (SR). }
    \label{viadapter}
    % \vspace{-10pt} % 缩短表标题与表格的距离
    \begin{tabular}{c|c|cc}
        \toprule
        Visual Adapter  & Params(M) & PR & SR \\
        \midrule
        - & 1.9 & 0.659 & 0.518 \\
        Spatial & 2.1 & 0.690 & 0.544 \\
        Spatial+Channel & 2.6 & 0.698 & 0.549 \\
        Spatial+Channel+Frequency & 3.8 & \textbf{0.718} & \textbf{0.565} \\
        \bottomrule
    \end{tabular}
    % \vspace{-15pt} % 缩短表格与正文的距离
\end{table}
To further validate the effectiveness of the proposed visual adapter, we conduct a comprehensive ablation study to investigate the contributions of its internal modules. The experimental results on the LasHeR dataset are presented in Table~\ref{viadapter}. We decompose the Visual Adapter into three components: spatial, channel, and frequency modules.
The first row displays baseline performance without the Visual Adapter, achieving PR (Precision Rate) and SR (Success Rate) of 0.659 and 0.518, respectively. When incorporating spatial information processing (second row), both metrics increased by 3.1\% and 2.6\%. The subsequent integration of channel information (third row) yielded additional improvements of 0.8\% and 0.5\%. The complete Visual Adapter (final row), incorporating all three modules, further enhanced performance by 2\% and 1.6\% in PR and SR, respectively.
The experiments conclusively validate the effectiveness of each component in the Visual Adapter and demonstrate the necessity of integrating frequency information.

\emph{3) Effectiveness of Memory Adapter: }
\begin{table}
    \centering
    % \vspace{-10pt}
    % \renewcommand{\arraystretch}{0.9} % 调整行间距
% \normalsize
    \small
    \setlength{\tabcolsep}{7pt}
    \caption{Effectiveness of memory adapter on the LasHeR dataset, the performance is evaluated in terms of precision rate (PR) and success rate (SR). }
    \label{memadapter}
    % \vspace{-10pt} % 缩短表标题与表格的距离
    \begin{tabular}{c|cc}
        \toprule
        Memory Adapter  & PR & SR \\
        \midrule
        Propagate to adjacent  & 0.659 & 0.518 \\
        Short Memory & 0.672 & 0.527 \\
        Short+Long Memory & 0.680 & 0.537 \\
        Short+Long+Permanent Memory & \textbf{0.689} & \textbf{0.545} \\
        \bottomrule
    \end{tabular}
    % \vspace{-15pt} % 缩短表格与正文的距离
\end{table}
To further investigate the effectiveness of the Memory Adapter, the experimental results on LasHeR are shown in Table~\ref{memadapter}. The baseline method, employing ODTrack's adjacent-frame propagation approach, achieves PR and SR scores of 0.659 and 0.518, respectively. Incorporating short-term memory improves these metrics by 1.3\% and 0.9\%, while subsequent integration of long-term memory yields additional gains of 0.8\% and 1.0\%. The complete Memory Adapter ultimately attains PR and SR scores of 0.689 and 0.545. These results demonstrate that our memory adapter more effectively utilizes temporal information compared to adjacent-frame propagation methods, while also confirming the rationality of our design.

\subsection{Parameter Analysis.}
% \begin{table}
%     \centering
%     % \normalsize
%     \small
%     \setlength{\tabcolsep}{12pt}
%     \caption{Parameter Analysis on the LasHeR dataset, the performance is evaluated in terms of precision rate (PR) and success rate (SR). }
%     \label{memadapter}
%     \begin{tabular}{c|cc}
%         \toprule
%         Memory Adapter  & PR & SR \\
%         \midrule
%         w./o. Memory & 0.718 & 0.565 \\
%         Retrieve + 888 size & 0.722 & 0.568 \\
%         Retrieve + 333 size & 0.720 & 0.567 \\
%         Mean + 883 size & 0.721 & 0.568 \\
%         \textbf{Retrieve + 883 size (Ours)} & \textbf{0.726} & \textbf{0.571} \\
%         \bottomrule
%     \end{tabular}
% \end{table}
\begin{table}[h]
  \centering
  \small
  \setlength{\tabcolsep}{8pt}
  \caption{Memory configuration study on LasHeR.  “S/L/P” denotes
           the number of tokens in \textbf{S}hort-, \textbf{L}ong- and
           \textbf{P}ermanent-term banks, respectively.}
  \label{tab:mem_cfg}
  \begin{tabular}{c|c|cc}
    \toprule
    Settings & Tokens (S/L/P) & PR & SR \\
    \midrule
    –                & 0/0/0   & 0.718 & 0.565 \\
    Retrieval        & 8/8/8   & 0.722 & 0.568 \\
    Retrieval        & 8/3/3   & 0.720 & 0.567 \\
    Mean             & 8/8/3   & 0.721 & 0.568 \\
    \textbf{Retrieval} & \textbf{8/8/3} & \textbf{0.726} & \textbf{0.571} \\
    \bottomrule
  \end{tabular}
\end{table}
To investigate how memory size and read-out strategy affect tracking performance, we conduct a parameter study on the LasHeR dataset, as shown in Table~\ref{tab:mem_cfg}. Starting from the strong visual-adapter baseline that has no memory at all (PR = 71.8\%, SR = 56.5\%), we first enable the memory module while keeping a large long-term bank of 8/8/8 tokens and our similarity-based retrieval. This configuration boosts performance to 72.2\% and 56.8\%, showing that simply giving the model more historical snapshots already helps it recover appearance information lost during occlusion. Shrinking the bank to 3/3/3 tokens, while still using similarity retrieval, reduces the gain to 72.0\% and 56.7\%, which implies that an undersized memory cannot fully capture the appearance diversity of the long LasHeR sequences.

Next, we fix the memory size to 8/8/3 but replace similarity-based retrieval with a naïve arithmetic mean across stored tokens. This yields 72.1\% (PR) and 56.8\% (SR), nearly identical to the large-bank setup, revealing that capacity alone is insufficient—without a mechanism to focus on relevant entries, crucial details are diluted. Finally, using both the 8/8/3 memory configuration and similarity-based retrieval produces our best results. This configuration balances capacity and selectivity, large enough to remember long-term appearance changes yet compact enough to avoid redundant noise, while the retrieval step ensures that only the most pertinent memories influence the current template.
\begin{figure}[t]
\centering
     \includegraphics[width=0.78\textwidth]{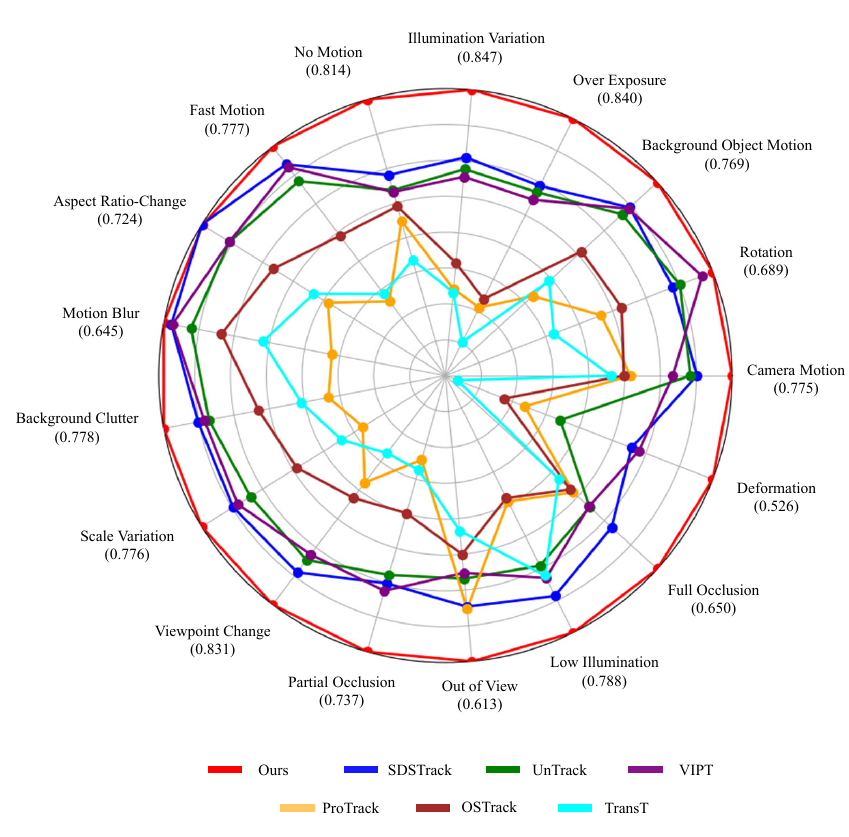}   
  \caption{Precision scores of different attributes on the VisEvent.} 
  \label{fig:radar}  
\end{figure}

\subsection{Attribute Analysis.} 
% We perform an analysis of various challenging attributes on VisEvent including full occlusion, overexposure, \emph{etc.} As in Figure~\ref{fig:radar}. Our method outperforms all existing approaches in these challenging scenarios. This improvement is attributed to the visual adapter, which selectively leverages multi-modal information, thereby enhancing robustness and adaptability in complex environments.

We perform an analysis of various challenging attributes on VisEvent including Illumination Variation, Over Exposure, Background Object Motion, Rotation, Camera Motion, Deformation, Full Occlusion, Low Illumination, Out of View, Partial Occlusion, Viewpoint Change, Scale Variation, Background Clutter, Motion Blur, Aspect Ratio-Change, Fast Motion, and No Motion. As in Figure~\ref{fig:radar}.
The Figure clearly shows that our tracker encloses a consistently larger area than all competing methods, indicating broad improvements rather than isolated wins on a few attributes.
The gain is most pronounced on full occlusion and overexposure, where the RGB stream is unreliable but the event modality still captures edge activity; the visual adapter dynamically up-weights the latter, sustaining stable localization after re-appearance.
We also observe non-trivial margins under fast motion and low illumination. This suggests that the frequency-aware recalibration mitigates motion blur and intensity saturation by selectively emphasizing high-frequency cues from events while suppressing noisy low-frequency components in RGB.
This improvement is attributed to the visual adapter, which selectively leverages multi-modal information, thereby enhancing robustness and adaptability in complex environments.

% To validate the effectiveness of each component of the proposed method, we conducted ablation experiments on the LasHeR dataset. The results are shown in the table~\ref{tableaba}.

\subsection{Comparison of Tunable Parameter.} 
To evaluate the impact of different components on the tunable parameter and model complexity, we conduct a comparison of tunable parameters on the LasHeR dataset. As shown in Table~\ref{tabletun}, the first row represents the baseline model, where all components except the patch-embedding layer are frozen, resulting in a total of 93.4M parameters, with 1.9M being tunable. The second row shows the result of making all components of the baseline tunable, which improves the tracking performance by 4.3\% in precision rate and 3.7\% in success rates; however, the number of tunable parameters rises to 93.4M. In the third row, the proposed visual adapter is added to the baseline and made learnable, adding only 1.9M tunable parameters, while improving the tracking performance by 5.9\% in precision rate and 4.7\% in success rate, compared to the baseline. The last row shows the final version of the proposed method, which has 98.9M model parameters and 7.3M of them are tunable. In addition, our method introduces about 1 GFlops computational cost more than the baseline.
% We analyze tunable parameters and model complexity on LasHeR (Table~\ref{tabletun}). The baseline (Row 1) freezes all components except patch embedding, using 93.4M total parameters (1.9M tunable). Fully tuning the baseline (Row 2) improves precision and success rates by 4.3\% and 3.7\% but increases tunable parameters to 93.4M. Adding our learnable visual adapter to the baseline (Row 3) introduces only 1.9M additional tunable parameters while boosting precision by 5.9\% and success rate by 4.7\%. The final method (Row 4) achieves 98.9M total parameters (7.3M tunable), balancing efficiency and performance.

\subsection{Comparison of Visual Adapters.}
% We conduct comparison experiments between our proposed visual adapter and the most commonly used adapter Fovea fusion, which has been widely used in many prompt-learning methods~\cite{VIPT,seqtrackv2,tatrack} and fine-tune it using the same way as the adapter we proposed. As shown in Table~\ref{tableadapter}, our tracker outperforms Fovea fusion in all the multi-modal tracking datasets. Notably, our method requires only 1.7M additional tunable parameters compared to Fovea fusion, highlighting its efficiency and effectiveness.
We conduct comparison experiments between our proposed visual adapter and the widely adopted Fovea fusion adapter~\cite{VIPT,seqtrackv2,tatrack}, fine-tuning both under identical settings. The “–” row in Table~\ref{tableadapter} represents a strong baseline without any dedicated adapter. Introducing the Fovea fusion adapter improves tracking performance on the LasHeR dataset, with precision and success rates increasing from 65.9\% and 51.8\% to 69.0\% and 54.4\%, respectively. Similar trends are observed on DepthTrack and VisEvent, where Fovea yields modest but consistent gains, indicating that spatial re-weighting of features contributes positively to performance. Replacing Fovea with our frequency-aware visual adapter leads to further improvements, achieving 71.8\% precision and 56.5\% success on LasHeR. This represents absolute gains of +5.9\% (PR) and +4.7\% (SR) over the baseline, and +2.8\% / +2.1\% over Fovea. A similar trend is seen on DepthTrack, where the F-score improves from 59.8\% (no adapter) and 60.4\% (Fovea) to 61.3\%. On VisEvent, precision rises from 78.4\% (baseline) and 79.0\% (Fovea) to 79.5\%.
Although our module introduces 3.8 M tunable parameters in total, this is only 1.7 M more than the Fovea variant and remains lightweight compared with full fine-tuning of the backbone. The extra capacity is used to model frequency, domain interactions across RGB and auxiliary modalities, which explains why the gain is most pronounced on LasHeR, where motion blur and illumination variation make high-frequency cues especially informative, while still yielding steady improvements on the other two datasets. In sum, Table~\ref{tableadapter} confirms that our adapter offers the best accuracy–efficiency trade-off among the tested designs, outperforming both the plain baseline and the strong Fovea fusion across all metrics without incurring a prohibitive parameter cost.
% In this section, we conduct comparison experiments between the multi-modal fusion adapter we proposed and the commonly used Fovea fusion adapter, which has been widely applied in various prompt-learning methods~\cite{VIPT,seqtrackv2}. The results, shown in Table~\ref{tableadapter}, indicate that our proposed method outperforms Fovea fusion across all multi-modal tracking datasets. Moreover, in terms of tunable parameters, our method has only 1.7M more than Fovea fusion, demonstrating that it is both efficient and effective.
\begin{figure*}[t]
\centering
  \includegraphics[width=0.95\textwidth]{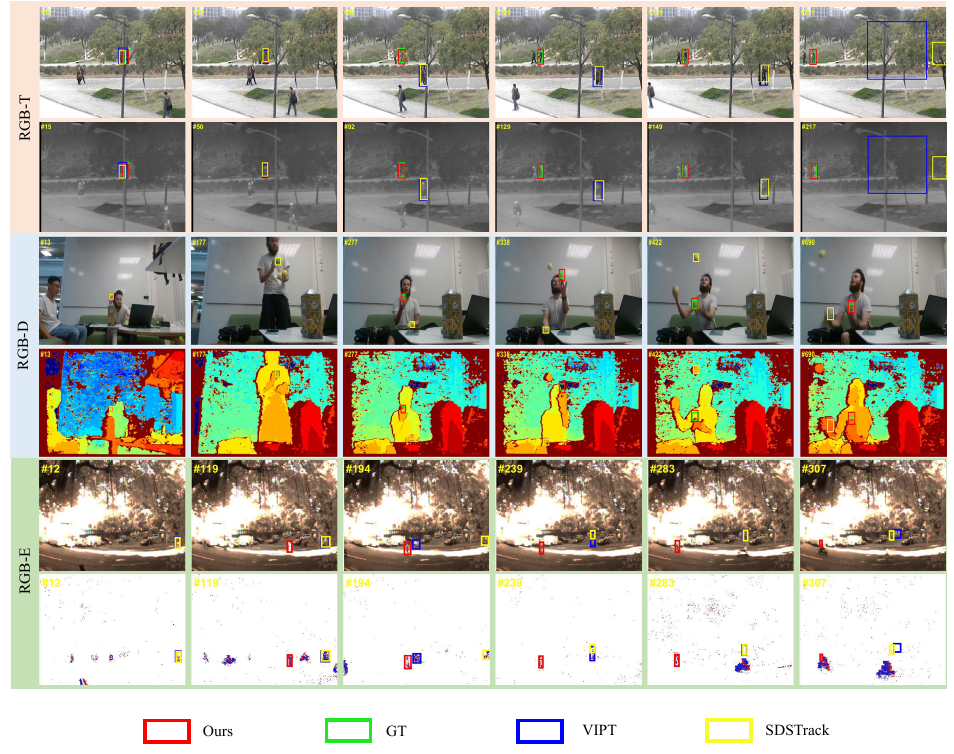}   
  \caption{Qualitative comparison of our method with ViPT and SDSTrack on RGB-T, RGB-D, and RGB-E tracking benchmarks.} 
  \label{fig:visual}  
\end{figure*}
\begin{figure}[t]
\centering
  \includegraphics[width=0.95\textwidth]{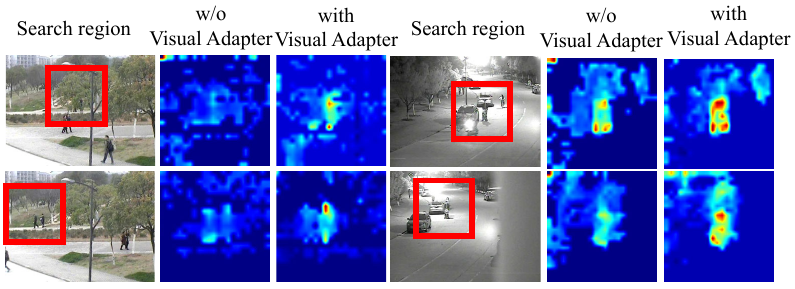}   
  \caption{Feature visualization before and after applying the visual adapter.} 
  \label{fig:featmap}  
  % \vspace{-15pt}
\end{figure}
\begin{figure}[t]
\centering
  \includegraphics[width=0.95\textwidth]{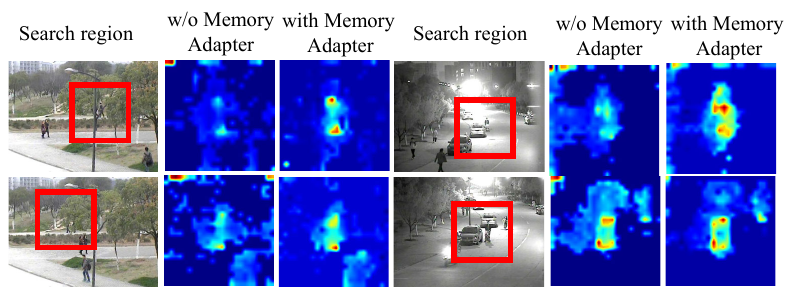}   
  \caption{Feature visualization before and after applying the memory adapter.} 
  \label{fig:memmap}  
  % \vspace{-15pt}
\end{figure}
\subsection{Qualitative Comparison and Feature Analysis.}
\emph{1) Qualitative Comparison with SOTA Methods: }
% We provide the visualization results of the proposed method compared with SOTA methods, as shown in Figure~\ref{fig:visual}. With the proposed visual adapter, the method generates more unambiguous and discriminative responses in complex scenarios, such as background clutter and similar objects. Additionally, the memory adapter assists the model in locating the object during occlusions. 
To intuitively demonstrate the effectiveness of our approach, we visualize tracking responses alongside two strong SOTA baselines, ViPT and SDSTrack, in Figure~\ref{fig:visual}.

The top row shows a representative RGB-T sequence with severe occlusions and multiple visually similar distractors. While both baselines quickly drift off the target, our tracker consistently maintains focus, benefiting from the memory adapter’s ability to preserve a reliable temporal trajectory.
The middle row presents a fast-motion RGB-D sequence containing several similar instances. Despite rapid camera and object movement, our method robustly localizes the correct target, highlighting its adaptability to depth-aided motion scenarios.
The bottom row illustrates a challenging RGB-E case with cluttered, low-resolution RGB frames and clean, high-temporal-resolution event data. By effectively fusing these complementary modalities, the visual adapter generates sharp and unambiguous response maps, enabling accurate tracking even under complex background interference.

\emph{2) Feature Representation Analysis: }
To provide a more comprehensive visual analysis, we further inspect the feature visualization produced by both the visual and memory adapters. 
For the visual adapter, Figure \ref{fig:featmap} arranges three columns: the left column shows the raw search region, the middle column depicts the feature map generated by the baseline without our adapter, and the right column presents the features after using the adapter. Once the adapter is enabled, the target contour becomes crisp and the clutter in the background is markedly suppressed, confirming that frequency-aware recalibration strengthens discriminative cues while filtering out irrelevant noise. 
The memory adapter cannot be plotted directly because it stores one-dimensional vectors, so we instead compare the target‐region feature maps before and after the memory adapter in Figure \ref{fig:memmap}. After memory aggregation, the foreground activations are noticeably sharper and the separation from the background is clearer, illustrating that temporal information preserved in memory further refines spatial representations and stabilizes tracking.

\section{Conclusion}
% In this paper, we proposed a novel prompt-learning-based tracker that combines visual and memory adapters for boosting multi-modal tracking performance. Our visual adapter effectively integrated auxiliary modalities into the RGB modality by learning robust and adaptable prompts that capture frequency characteristics. Additionally, our memory adapter was designed to store and retrieve global temporal context, ensuring that the most reliable temporal cues are propagated through sequential frames. Extensive experiments demonstrate that our tracker achieves state-of-the-art performance in all multi-modal tracking benchmarks.

In this paper, we proposed a novel prompt-learning framework that learns frequency- and memory-aware prompts for multi-modal object tracking. The frequency-guided visual adapter adaptively fuses auxiliary modalities with RGB by exploiting complementary spatial, channel, and frequency information, while the multi-level memory adapter stores and retrieves temporal cues to ensure consistent long-range propagation across frames. This dual-adapter design enables efficient and robust prompt learning on frozen trackers. Extensive experiments on RGB-T, RGB-D, and RGB-E benchmarks demonstrate that our method achieves state-of-the-art performance, significantly outperforming both fully fine-tuned and adapter-based baselines.

\section*{Acknowledgment}
This work was supported by the National Natural Science Foundation of China (62072232, 62576098), the Key R\&D Project of Jiangsu Province (BE2022138), the Fundamental Research Funds for the Central Universities (021714380026), the program B for Outstanding Ph.D. candidates of Nanjing University, and the Collaborative Innovation Center of Novel Software Technology and Industrialization. \emph{(Boyue Xu and Ruichao Hou contributed equally to this work.)}
\bibliographystyle{elsarticle-num} 
\bibliography{ref}




%% For numbered reference style

  % Leslie Lamport,
  % \textit{\LaTeX: a document preparation system},
  % Addison Wesley, Massachusetts,
  % 2nd edition,
  % 1994.

% \end{thebibliography}
\end{document}